\documentclass{article}

\usepackage[final]{nips_2017}
\usepackage{wrapfig}
\usepackage{graphicx}
\graphicspath{ {images/} }
\usepackage{natbib}
\setcitestyle{square,numbers}

\usepackage[utf8]{inputenc} % allow utf-8 input
\usepackage[T1]{fontenc}    % use 8-bit T1 fonts
\usepackage{hyperref}       % hyperlinks
\usepackage{url}            % simple URL typesetting
\usepackage{booktabs}       % professional-quality tables
\usepackage{amsfonts}       % blackboard math symbols
\usepackage{nicefrac}       % compact symbols for 1/2, etc.
\usepackage{microtype}      % microtypography

\title{ParaNet - Using Dense Blocks for Early Inference}

\author{
  Joseph Chuang \\
  Cornell University \\
  \And
  Eric Tsai \\
  Cornell University \\
  \And 
  Kevin Huang \\
  Cornell University\\
  \And 
  Jay Fetter \\
  Cornell University
}

\begin{document}

\maketitle

\begin{abstract}
DenseNets have been shown to be a competitive model among recent convolutional network architectures. These networks utilize Dense Blocks, which are groups of densely connected layers where the output of a hidden layer is fed in as the input of every other layer following it. In this paper, we aim to improve certain aspects of DenseNet, especially when it comes to practicality. We introduce ParaNet, a new architecture that constructs three pipelines which allow for early inference. We additionally introduce a cascading mechanism such that different pipelines are able to share parameters, as well as logit matching between the outputs of the pipelines. 
We separately evaluate each of the newly introduced mechanisms of ParaNet, then evaluate our proposed architecture on CIFAR-100.
\end{abstract}

\section{Introduction}
Convolutional Neural Networks (CNNs) are the state-of-the-art class of algorithms currently available for image classification. A plethora of architectures have been proposed and benchmarked on various datasets, each introducing a new technique for constructing deep networks (for example, ResNet introduced residual connections, and DenseNets introduced dense connections). Each iteration improved upon the last, and performance have improved up to such a point that the 2017 ImageNet competition will be the last one held, as the dataset is on the verge of being a "solved problem".

The next step for deep learning is the ability to deploy these systems into an application context. Model compression has been an active area of research given that in practical settings, it's not always feasible to conduct inference with a large model. For example, given a small mobile device such as a Raspberry Pi with finite computational resources, we would want to minimize the number of floating point operations that a model would need to compute in order to arrive at its prediction. This would allow the model to make real-time predictions on systems with limited computational resources.

\section{Related Work}

Recent architectures have popularized the usage of skip connections, essentially linking up earlier layers closer to the output. DenseNets focused on using dense blocks as an extension of skip connections, where every layer in a dense block feeds into all subsequent blocks. The authors state that the benefits of using dense connections include alleviating the vanishing gradient problem, efficient feature reuse and propagation, as well as reduced number of overall parameters.
\cite{densenet}

Multi-Scale Dense Nets, or MSDNet, is another architecture that also takes advantage of dense connections. However, instead of just using Dense Blocks to improve training time and overall prediction accuracy, the goal of MSDNet was improving the performance on anytime and budgeted classification tasks. The MSDNet is able to beat the current state-of-the-art models in terms of accuracy based on the number of FLOPS. MSDNet does so by early exiting depth-wise, such that confident samples are inferred quickly, whereas less confident samples are sent deeper within the network. Also, dense connections throughout different layers in such a network allow for feature sharing and reuse.
\cite{msdnet}

Distillation is a technique where the performance and characteristics of a large model are transferred to a smaller model. A simple way to do so is to perform logit matching, where the inference outputs of a large model are used as training targets for a smaller model. In \cite{distill}, distillation is defined as similar to logit matching, but with temperature scaling on the softmax targets as well as mixing with the original hard, one-hot targets.

\section{ParaNet3}
\begin{figure}
\begin{center}
\includegraphics[width=.6\textwidth]{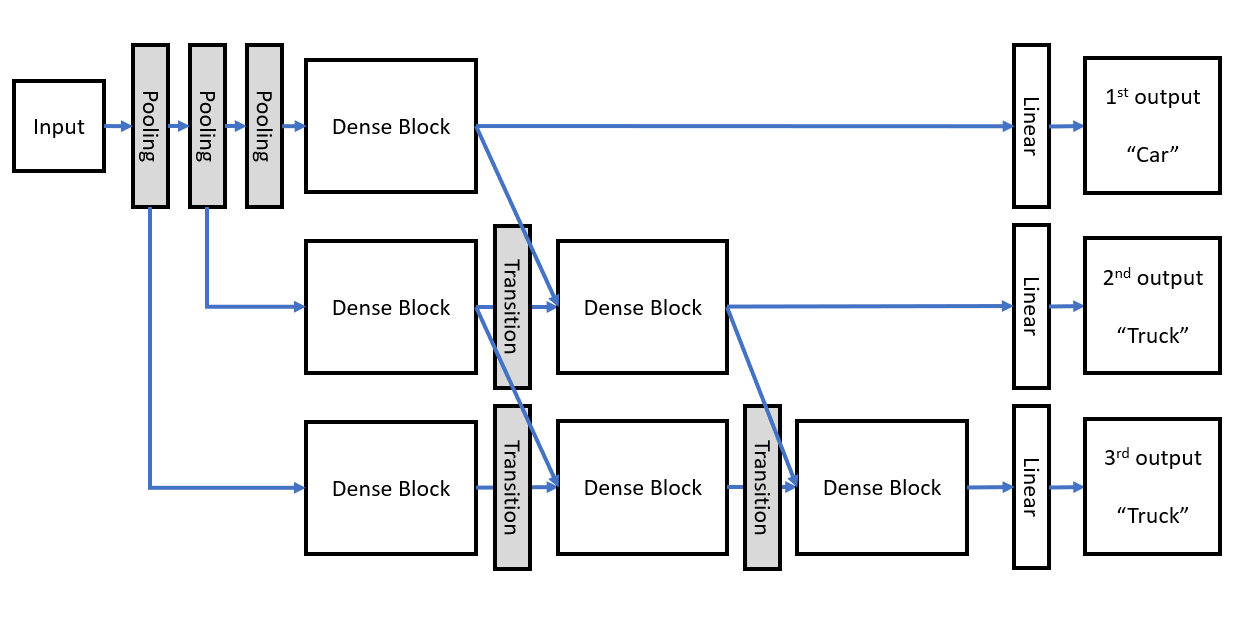}
\end{center}
\caption{ParaNet Architecture}
\end{figure}

Our new architecture, ParaNet3, employs several of the aforementioned ideas in a proof-of-concept model. The general idea is that we have 3 pipelines - pipeline 1 on top, 2 in the middle, and 3 at the bottom. Each pipeline is responsible for its own set of predictions, which are the outputs of a final fully connected layer. First, we directly pool the input to feed into each pipeline, such that inference on the smaller pipeline should be significantly faster. Then, we reuse filters and feature maps throughout the network such that later, heavier pipelines can benefit from earlier computations. Finally, we can opt for training with logit matching between pipelines for benefits conferred by distillation.

\subsection{Pooling}
We directly pool at the beginning; each pooling layer reduces each dimension by half. For each transition block, a similar pooling process takes place. Therefore, all Dense Blocks at the end of each pipeline are working with input volumes that have been pooled three times, and vice versa for Dense Blocks that are second to last in each pipeline with input volumes that are pooled twice, etc.

\subsection{Dense Block}
Our architecture heavily relies on the usage of Dense Blocks. Each Dense Block is identical in terms of the number of layers. We follow the setup for Dense Block as per the original paper \cite{densenet}. Further details about specific hyper-parameters for the Dense Blocks are provided in the Experimental Section.

\subsection{Cascading}
Each of the Dense Blocks may feed into multiple destinations. First, Blocks may feed into a Fully Connected/Linear layer, where a softmax output is computed. However, Blocks may also feed into the next Block in the same pipeline, or the next Block in the next pipeline. The former case requires a Transition Block that pools the output volume, such that the working feature maps are decreasing over depth. However, given the extra pooling performed for earlier pipelines, feeding into the next block in the next pipeline does not require pooling - this part of the network is referred to as "cascading". These two inputs from the previous and current pipeline are simply concatenated as a singular input.

Alternatively, one may construct a similar network without the "cascading", where none of the pipelines share any parameters. We denote ParaNet3 that includes cascading with the prefix PN3, and ParaNet3 that does not include cascading with PN3cut.

\subsection{Logit Matching}
As described in Related Work, Logit Matching is a common practice for model compression. While training a small pipeline, we match its output to the logit of the large pipeline. \cite{distill} shows that this compression technique allows the compressed model to compress the knowledge of the large model into a small model. 

Given that we have 3 pipelines, we denote our logit matching configurations with three characters, which form the suffix for our model labels. The use of a number represents logit matching towards a certain pipeline; the use of the 'd' character represents matching towards the "true" logit targets; the use of the 'x' character represents that a specific pipeline is not matched towards a particular target, and therefore the pipeline is left untrained. The first digit represents the target of the first pipeline, and vice versa for the second and third. 

For example, PN3-dxx represents a model with cascading and only the first pipeline matched to the true logits, whereas PN3-33d represents a model where the first two pipelines are matched to the third, and the third matched to true logits.

\subsection{Early Inference}
ParaNet is composed of multiple pipelines which have more dense blocks in the larger pipeline. Because each pipeline can make the prediction in the inference phase, inferencing a data point could exit at the small pipeline when the prediction confidence is high enough. In other words, ParaNet can use the small pipeline to conduct inference on the easy data point and the large pipeline to inference the hard data point. Early inference decreases the computations needed to achieve good enough inference accuracy.

\section{Experiments}
We aim to test the effect of each of our techniques - logit matching, cascading, and early inference - on Paranet. All tests are performed by training and validating on CIFAR-100.

For training, we fix the sizes of the networks to 100 total layers with 12 growth. 
We employ a drop-based learning rate of schedule, set to start at 0.1, decrease to 0.01 at 50 epochs, and finally 0.001 at 100 epochs. We train for a total of 130 epochs. We do not reduce the number of filter maps at each Transition Block. We use bottleneck layers as described in \cite{densenet} for all Dense Blocks. We also perform batch normalization before the fully connected layers.

\subsection{Cascading}
This experiment answers whether cascading the output of the upper pipeline to the lower pipeline helps the prediction of the cascaded pipeline. We compare the accuracy of ParaNet architecture with and without cascading the pipelines. Specifically, we compare PN3cut-ddd and PN3-ddd to test the effect of the cascading connections.

\begin{table}[h]
\begin{center}
\begin{tabular}{llr}
\hline
 & PN3cut-ddd & PN3-ddd		\\
\hline
Pipeline 1 & \boldmath$50.39 \pm 0.248$ & $48.23$\\
Pipeline 2 & $65.584 \pm 0.132$ & $65.63$ \\
Pipeline 3 & $74.582 \pm 0.234$ & \boldmath$76.24$\\
\hline
\end{tabular}
\caption{Logit Matching results, by Validation Accuracy}
\end{center}
\end{table}

Table 1 shows that the accuracy of pipeline 1 and pipeline 2 are not improved with cascading. This might be caused by the confusing objective that train the pipeline to both make the prediction and help the prediction of the next pipeline at the same time. Pipeline 3 does have a 2\% improvement, probably due to larger network contributing to its prediction.

\subsection{Logit Matching}
This experiment verifies that matching the output of the current pipeline to the logit of the next pipeline can improve the accuracy of the current pipeline. We compare the accuracy of the second pipeline with and without logit matching to the third pipeline. Specifically, we compare PN3-xdx and PN3-x3d to test the effect of logit matching on pipeline 2. Table 2 shows that logit matching significantly improves the second pipeline.

\begin{table}[h]
\begin{center}
\begin{tabular}{llr}
\hline
Network		& Validation Accuracy\\
\hline
PN3-xdx		& $65.674 \pm 0.346$		\\
PN3-x3d 	& \boldmath$67.024 \pm 0.311$	\\
\hline
\end{tabular}
\caption{Logit Matching results}
\end{center}
\end{table}

\subsection{Early Inference}
This experiment shows that ParaNet architecture contributes higher accuracy with the computational budget. We calculate FLOP counts for each of the pipelines in PN3-ddd and graph their respective accuracies.

\begin{figure}[h!]
\centering
\includegraphics[width=.4\textwidth]{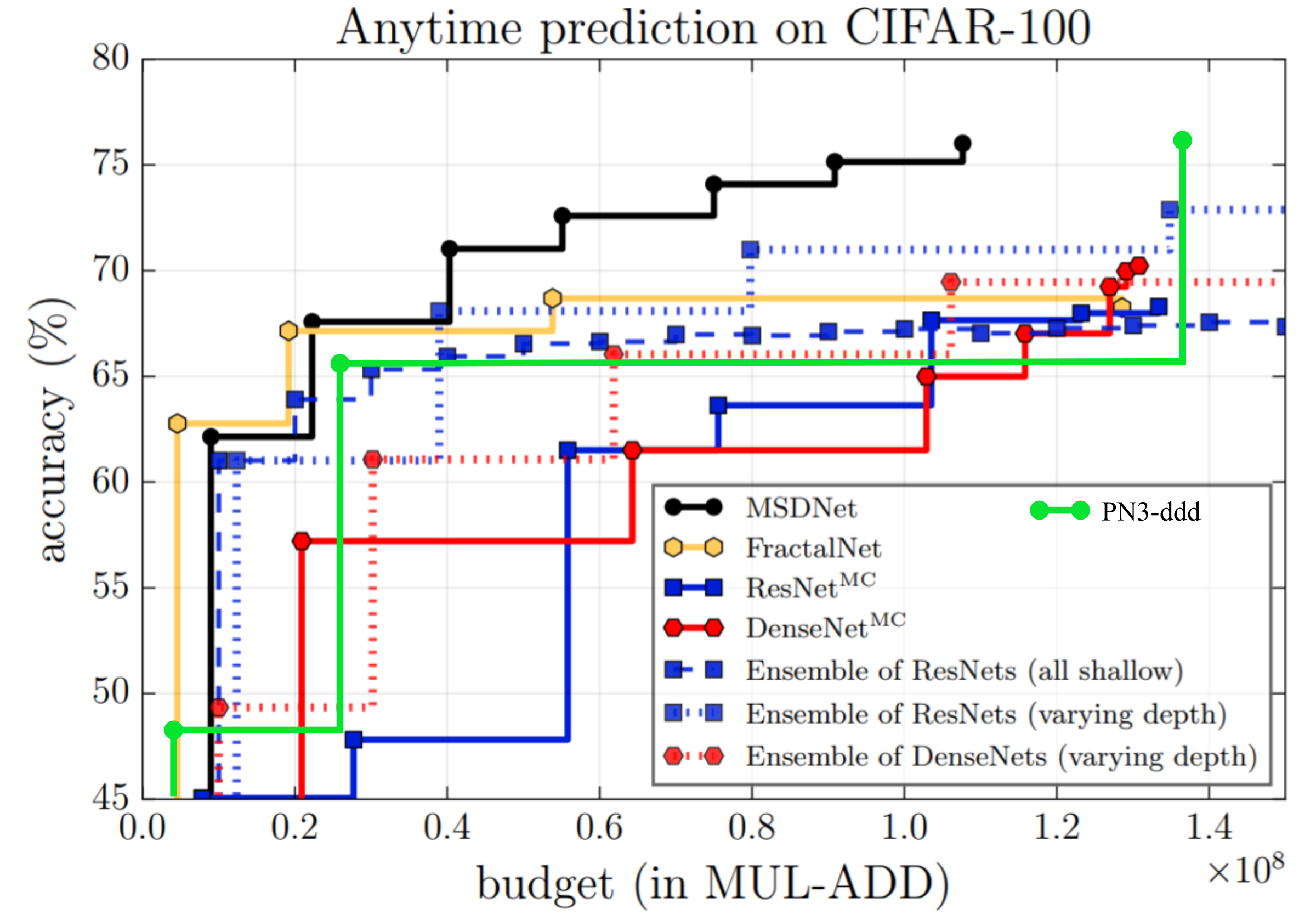}
\caption{Anytime Inference results}
\end{figure}

The result in Figure 2 shows that ParaNet outperforms some ensembled methods, yet falls short of current state-of-the-art such as Fractal Nets and MSDNets.

\section{Conclusion}
Cascading does not seem to improve the predictions of our pipelines as we anticipated. This is most likely due to a confusion of objectives, where the first Dense Block is simultaneously being used for multiple pipeline outputs.

Logit matching significantly increases the accuracy of the second pipeline. This confirms the intuition behind logit matching and model distillation.

ParaNet has potential to be competitive with better tuning and perhaps some architecture modifications - especially with more exit points. In Figure 2, we clearly observe that FLOP counts for pipeline 2, 3 are too far apart for ParaNet to be effective at early inference.

\section{Future Work}
We would like to look into evaluating ParaNet based on Budgeted Batch classification, where the thresholds for early inference actually come into play. This may involve confidence calibration using temperature scaling as explored in \cite{calibration}, where simply rescaling logits before the softmax fixes issues of overconfidence in deep networks.

We would also like to test more variations of ParaNet. First, we should test more "cut" architectures that do not involve cascading. We could also add to the number of pipelines, or perhaps make each pipeline contain the same number of Dense Blocks. This is to maintain consistency and solve the gap between FLOP counts, especially between the second and third pipelines.

\bibliography{citations}

\end{document}